\documentclass[11pt,a4paper]{article} 
\pdfoutput=1
\usepackage[utf8]{inputenc} 
\usepackage{bm}
\usepackage{amsmath}
\usepackage{amsthm}
\usepackage{amssymb}
\usepackage{boldline}
\usepackage{float}
\usepackage{fancyhdr}
\usepackage{multirow}
\usepackage[ruled,vlined,linesnumbered,resetcount]{algorithm2e}

\usepackage{geometry} 
\usepackage{graphicx} 

\usepackage{booktabs} 
\usepackage{array} 
\usepackage{paralist} 
\usepackage{verbatim} 
\usepackage{subfig} 
\usepackage{sectsty}
\allsectionsfont{\sffamily\mdseries\upshape} 

\DeclareMathOperator*{\argmin}{arg\,min}

\title{Detecting Adversarial Examples in Convolutional Neural Networks}
\author{Stefanos Pertigkiozoglou \and Petros Maragos}
\date{School of Electrical and Computer Engineering\\ National Technical University of Athens, Athens, Greece \\  stefanospert@gmail.com, maragos@cs.ntua.gr} 

\begin{document}
\maketitle
\begin{abstract}
\par The great success of  convolutional neural networks  has caused  a massive spread of the use of such models  in a large variety of Computer Vision applications. However, these models are vulnerable to certain inputs, the adversarial examples, which although are not easily perceived by humans, they can lead a neural network to produce faulty results. This paper focuses on the detection of adversarial examples, which are created for  convolutional neural networks that perform image classification. We propose three  methods for detecting possible adversarial examples and after we analyze and compare their performance, we combine their best aspects to develop an even more robust approach.
\par  The first proposed  method is based on the regularization of the feature vector that the neural network produces as  output. The second method detects adversarial examples by using  histograms, which are created from  the outputs of the hidden layers  of the neural network. These histograms create a feature vector which is used as  the input of an SVM classifier, which classifies the original input either as an adversarial  or as a real input. Finally, for the third method we introduce the concept of the residual image, which contains information about the parts of the input pattern that are ignored by the neural network. This method aims at the detection of possible adversarial examples, by using the residual image and reinforcing the parts of the input pattern that are ignored by the neural network. Each one of these methods has some novelties and by combining them we can further improve the detection results.
For the  proposed methods and their combination, we present the results of detecting adversarial examples on the MNIST dataset. The combination of the proposed methods offers some improvements over similar state of the art approaches.
\end{abstract}
\section{Introduction} \label{sec:1}
\indent
\par The use of  Deep Learning and deep neural networks has spread in a large variety of Computer Vision applications due to their increasingly effectiveness in solving many difficult visual tasks. Specifically a  Convolutional Neural Network, presented in \cite{LBH16}, is a deep learning model which is used extensively in image recognition. A  Convolutional Neural Network (CNN) consists of successive layers, where the network processes the input patterns in different scales. These multiple levels of representation remove the need for complex feature extraction, which transforms the raw data into a feature vector, because a CNN can accept the raw data as input and learn how to extract the important features internally in its  first few layers. In addition, these types of neural networks take advantage of the locality of the patterns in an image by using convolutional layers. 
\par However, neural networks and subsequently CNNs are vulnerable to certain inputs as shown in \cite{Sze+13}, the adversarial examples. These inputs, although are not easily perceived by humans can lead a CNN to produce faulty results. In the case of CNNs that are used in image classification, where the input is an image, we are also referring to adversarial examples as adversarial images. In this case, an original image,  which is classified correctly by a CNN, is slightly perturbed to produce the adversarial image. Despite the fact that the adversarial image seems similar to the original image, according to the perception of a human, it is classified into a different category. This means that a user can alter the output of the network by perturbing the input in a way that it is not detected by humans.
\par The existence of adversarial examples indicates that existing CNNs, although in some specific applications can achieve near human accuracy, they do not perceive the input in the same way as humans do. As a result, by studying adversarial examples we can improve the models used, in order to create models that are closer to the human perception, which we consider as the ideal solution. In fact, as shown  in \cite{GSS14}, by improving the behavior of a  neural network against adversarial examples we can also improve the accuracy of that network for real inputs. Also adversarial examples are a vulnerability that can be abused from a malicious user to influence the behavior of a system that uses a vulnerable neural network. For example a physical world adversarial example \cite{Eyk+18}, can alter the perception of a self-driving car that uses cameras to navigate through the urban environment.
\par There are two main approaches for combating adversarial examples. The first approach aims at making the neural network more robust against adversarial examples \cite{Pap+16}, \cite{ZCDL17}, by changing the network's architecture and the learning procedure. The second approach assumes that the neural network is already trained and tries to detect whether a new input is an adversarial example or it is a real input \cite{FCSG17}, \cite{Son+17}, \cite{MGFB17}, \cite{LIF17}, \cite{GWK17}.
\par  In this paper, we focus on the second approach and we propose methods that aim to detect  adversarial inputs. Specifically, we propose three different methods for detecting adversarial examples generated for a CNN that performs image classification. After we analyze and compare their performance, we propose different ways of  combining their best aspects to develop a more robust approach. The first method is based on the regularization of the feature vectors which are produced by the network. Using the regularized feature vectors  we retrain the last layer of the CNN similarly to the adversarial training proposed in \cite{GSS14}. We then can detect if a new input is an adversarial example  by comparing the output of the original network with the output of the retrained network.  The second method creates histograms using the absolute values of the outputs of the network's hidden layers. Then by combining these histograms, this method  creates a vector which is used by an SVM classifier to classify the input either as real or as adversarial. For the third method  we assume that in a neighborhood of the input space the CNN acts as an affine classifier. Using that assumption we introduce the concept of the residual image, which contains information about the parts of the input pattern that are ignored by the network. This information is then used to perturb the input image in order to detect whether this image is a real or an adversarial input . 
\par We use  these methods and their combination to detect adversarial examples generated for a LeNet \cite{LBBH98} network trained on the MNIST \cite{LBBH98}  dataset. The combination of the three proposed methods offers some improvements over similar state of the art approaches, on the detection of adversarial examples on the MNIST dataset.
\pagebreak
\section{ Notation amd Related Work} \label{sec:2}
\subsection{Notation} \label{sec:1.2}
\indent
\par The adversarial examples examined in this paper are generated for CNNs, which accept as input either a 2D signal of a grayscale image or a 3D signal of an RGB image. To simplify the notation, we use an image vector \(\bm{x}\in \mathbb{R}^m\) to denote an input image, where \(m\) is the total number of points of the discrete 2D or 3D signal. Each component of the image vector \(\bm{x}\)  corresponds to a certain point of the discrete 2D or 3D signal.
 Also, we refer to the \(i^{th}\) component of the image vector \(\bm{x}\) with \(x[i]\). This notation is also extended for the feature maps, which are the outputs of the hidden layers of the CNN.
\par When the image vector \(\bm{x}\) corresponds to a grayscale image, the operation corresponding to a discrete 2D convolution between the image and a kernel is done using a convolutional matrix that is multiplied with the image vector. 

\subsection{Generating Adversarial Examples} \label{sec:2.1}
\indent
\par In order to generate adversarial examples we assume that we have an already trained CNN that accepts as input an image and classifies it into one of N different categories. In particular the output of the CNN is a vector with N components, where the \(i^{th}\) component is the confidence of the network that the input belongs to the  \(i^{th}\) category. In addition, we have a set of test images (not used in training) which are used as starting points for the adversarial generation. We also refer to these test  images as real images.
\par For the trained network, let \(\bm{x}\) be the input of the network, \(f(\bm{x})\) be  the output of the network, \(k(\bm{x})\) be the category the input \(\bm{x}\)  is classified into  and \(J(f(\bm{x}),\ell)\) be the classification  error of the network with input \(\bm{x}\) and target category label \(\ell\in\{1,2,...,N\}\), where \(N\) is the total number of categories into which the network can classify the input. According to \cite{Sze+13}  if we have a real image \(\bm{x}\), which is classified into category with label \(\ell\), we can produce an adversarial image\mbox{ \(\bm{x}_\mathrm{adv}=\bm{x}+\bm{r}\)}, by solving the optimization problem
\begin{align} \label{eq:1}
\min_{\bm{r}} & \enskip c\lVert \bm{r} \rVert_2-J(f(\bm{x}+\bm{r}),\ell)   \\
\mathrm{s.t.: }& \enskip  k(\bm{x})\not=k(\bm{x}+\bm{r}) \label{eq:0} \nonumber
\end{align}
where \(c\) is a parameter that controls the \(L_2\) distance between the real image \(\bm{x}\) and the adversarial image \(\bm{x}_\mathrm{adv}\).
\par Instead of solving the optimization problem of  Equation (\ref{eq:1}), there are  many proposed methods \cite{RRB16},\cite{Dez+16},\cite{KGB17},\cite{CaWa17},\cite{Pap+16b}  that can produce  robust adversarial examples much faster. For the experiments on this paper we are using the Basic Iterative Method \cite{KGB17} and the DeepFool method \cite{Dez+16}. 

\subsubsection{Basic Iterative Method (BIM)}\label{sec:2.1.1}
One method proposed in \cite{GSS14}, as a faster alternative of solving the optimization problem of Equation (\ref{eq:1}), is the Fast Gradient Sign Method. In this method the adversarial image is produced by adding to the original image \(\bm{x}_0\), which is classified into the correct category \(\ell\), the vector \(\bm{r}\), where :
\begin{equation}
\bm{r}=\epsilon \cdot \mathrm{sign}(\nabla_x J(f(\bm{x}_0),\ell)) 
\end{equation}
As an extension of this method, \cite{KGB17} proposed the Basic Iterative Method (BIM), where the adversarial image is created by applying the fast gradient sign method several times with a smaller step \(a\), and also by clipping the result in each iteration in order to stay in a \(L_{\infty}\) \(\epsilon\)-neighbourhood of the original image \(\bm{x}_0\). This means that at iteration \(k\), the method generates an image \(\bm{x}_{k}\) where:
\begin{equation} \label{eq:2}
\bm{x}_{k}=\mathrm{Clip}_{\bm{x}_0,\epsilon}\{\bm{x}_{(k-1)}+a \cdot \mathrm{sign}(\nabla_x J(f(\bm{x}_{(k-1)}),\ell))\}, \quad \bm{x}_{(0)}=\bm{x}_0
\end{equation}
and the \(i^{th}\)  pixel of  \(\mathrm{Clip}_{\bm{x},\epsilon}(\bm{x}')\) is computed as follows:
\begin{equation} \label{eq:3}
\mathrm{Clip}_{\bm{x},\epsilon}(\bm{x}')[i]=\min{\{u_\mathrm{max}, x[i]+\epsilon,\max{\{u_\mathrm{min},x[i]-\epsilon,x'[i]\}}\}}
\end{equation}
where \(x[i],x'[i]\) are the \(i^{th}\) pixels of \(\bm{x},\bm{x}'\) respectively, and \(u_\mathrm{min},u_\mathrm{max}\) are the minimum and maximum  values allowed for the input.
\par The BIM method terminates at iteration \(k\), when it finds an adversarial image \mbox{ \(\bm{x}_\mathrm{adv}=\bm{x}_{k}\)} that it is classified into a category that it is different from the original category \(\ell\).

\subsubsection{DeepFool} \label{sec:2.1.2}
\par This method proposed in \cite{Dez+16} is an iterative method that is based on the linearization of the classifier at each step.
\par Let \(f_\ell(\bm{x})\) be the output of the trained network for the category \(\ell\) when the  input is the image vector \(\bm{x}\). If we have a real image \(\bm{x}_0\) which is correctly classified into the category \(\ell_0\), then using the DeepFool method after \(k\) iterations we can compute the image \(\bm{x}_{k+1}\) as follows
\begin{equation}\label{eq:5}
\ell=\argmin_{\ell\neq \ell_0} \frac{|f_\ell(\bm{x}_k)-f_{\ell_0}(\bm{x}_k)|}{\lVert \nabla f_\ell(\bm{x}_k)-\nabla f_{\ell_0}(\bm{x}_k)\rVert_2}
\end{equation} 
\begin{equation}\label{eq:6}
\bm{w}_k=\nabla f_\ell(\bm{x}_k)-\nabla f_{\ell_0}(x_k)
\end{equation}
\begin{equation}\label{eq:7}
\bm{x}_{k+1}=\bm{x}_k+\frac{|f_\ell(\bm{x}_k)-f_{\ell_0}(\bm{x}_k)|}{\lVert \bm{w}_k \rVert_2^2}\bm{w}_k
\end{equation}
\par This method terminates when it finds an image \(\bm{x}_k\) that it is classified  by the neural network  into a category \(\ell\not=\ell_0\). 
\newcommand{\costumsa}{0.31\textwidth}
\newcommand{\costumsb}{0.31\textwidth}
\begin{figure}[h]
\captionsetup[subfigure]{labelformat=empty}
\makebox[\textwidth][c]{
\subfloat[][]{
\includegraphics[width=\costumsa,height=\costumsb]{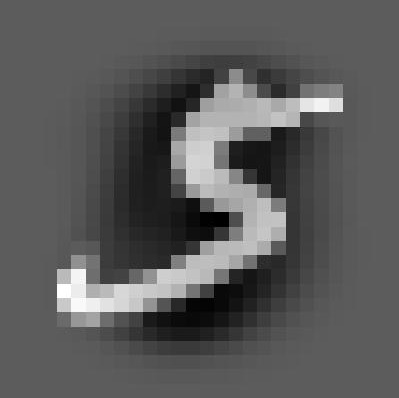}
\label{fig:1}
}
\subfloat[][]{
\includegraphics[width=\costumsa,height=\costumsb]{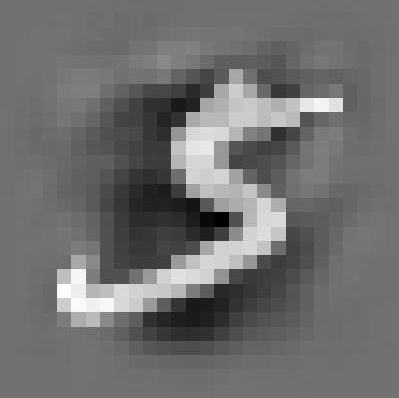}
\label{fig:2}
}
\subfloat[][]{
\includegraphics[width=\costumsa,height=\costumsb]{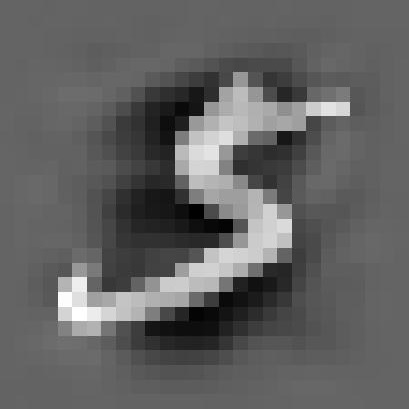}
\label{fig:3}
} 
}
\\[-3ex]
\makebox[\textwidth][c]{
\subfloat[][(a)]{
\includegraphics[width=\costumsa,height=\costumsb]{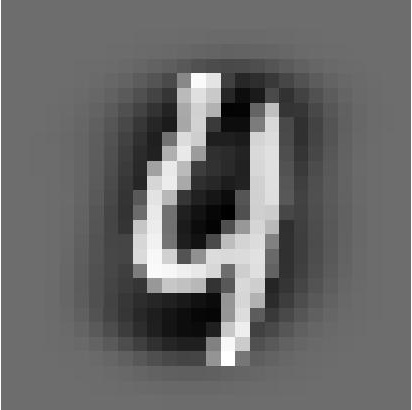}
\label{fig:4}
}
\subfloat[][(b)]{
\includegraphics[width=\costumsa,height=\costumsb]{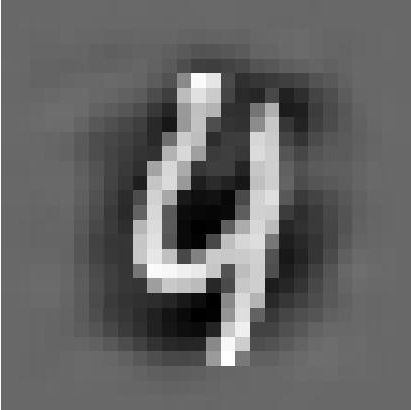}
\label{fig:5}
}
\subfloat[][(c)]{
\includegraphics[width=\costumsa,height=\costumsb]{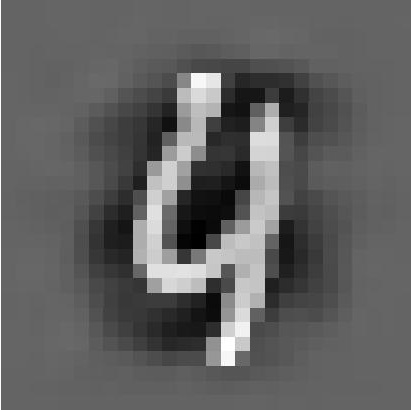}
\label{fig:6}
} 
}
\caption{\underline{Columns}: (a) Real input images, (b) Adversarial images produced with the BIM method, which are misclassified by the LeNet network, (c) Adversarial images produced with the DeepFool method, which are misclassified by the LeNet network}
\label{fig:101}

\end{figure}

\subsection{Detecting Adversarial Examples} \label{sec:2.2}
\indent
\par There is a large variety of different aprroaches that aim to detect possible adversarial inputs in a deep neural network. Many of these approaches are  based on the extraction of a certain metric, using the outputs of the neural network, which is then used to distinct between a real  input and an adversarial input.
 \par An example of two metrics that can be used for that distinction is proposed in \cite{FCSG17}. The first metric is based on the assumption that many adversarial generating methods produce adversarial examples  that are near the low dimensional submanifold where the real  inputs  lie (but not on it).  As a result \cite{FCSG17}  proposes a method to model the submanifolds of the real data using kernel density estimation in the feature space produced by the last hidden layer. Therefore an adversarial input will  produce a feature vector that is located in a region where the density estimate is lower than the density estimate for real  inputs. The second metric is another approach to identify low confidence regions of the input space. In particular,  given an input \(\bm{x}\), this metric computes  an estimation of the uncertainty of a deep Gaussian process \cite{GaGh15}, using the outputs of multiple DNNs with the same architecture that are trained in the same training set but using the dropout method. Using this metric we expect a higher uncertainty estimation when the input \(\bm{x}\) is an adversarial input.
\par Also a training procedure that can enhance the detection results when we use the kernel density estimation as a metric is proposed in \cite{PDDZ18}. This training procedure adds a regularization term called reverse cross-entropy. This term encourages the network to produce outputs which have high confidence for the correct category and confidence that is as uniform as possible for the other categories. 
\par Another method that aims at the detection of possible adversarial examples is proposed in \cite{Son+17}. This method uses the PixelCNN model \cite{SKCK17},  which is a trainable generative model. With this model  the likelihood \(p(\bm{x})\)   a new image \(\bm{x}\) is produced  by the model   can be easily computed. Based on the observation that adversarial inputs tend to be more unlikely to be produced by the model than real inputs, for a new image \(\bm{x}\) the value of \(p(\bm{x})\) can be used to detect whether this image is an adversarial example.
\par Apart from the methods that use a combination of metrics in order to detect possible adversarial examples, there are also proposed detection methods \cite{MGFB17},\cite{LIF17},\cite{GWK17} which train secondary neural networks that act as classifiers, which classify an input image either as an adversarial example or as a real input.
\section{ Proposed Methods for Detecting Adversarial Examples} \label{sec:3}
\subsection{ Regularization for detecting Adversarial Examples} \label{sec:3.1}
\indent
\par This method is based on the regularization of the feature vector that the neural network produces. For the regularization, we use the method  of nonlocal discrete regularization on weighted graphs proposed in \cite{EZB08}. Due to the fact that the outputs of the  layers closest to the final output of a CNN act as feature vectors of the input, we can use the outputs of one of these layers as the feature vector  of the input image.
\subsubsection{Feature Vector Regularization}\label{sec:3.1.1}
\indent
\par Let \(h(\bm{x})\) be the output of the second to  last layer of the CNN, when the input is the image \(\bm{x}\). For this method we are using \(h(\bm{x})\) as the feature vector of the image \(\bm{x}\), which is extracted by the neural network.
 To perform the regularization of the feature vector we create a weighted graph from a set of input images, where each vertex represents one of the input images. In addition, we define a function \(g\) on the vertices of the graph, where if the vertex \(u\) represents the input image \(\bm{x}\) with feature vector \(h(\bm{x})\) then \(g(u)=h(\bm{x})\). 
Let \(w(u,v)\) be the weight of the edge that connects the vertices \(u,v\). The norm of the gradient of \(g\) at a vertex \(u\) is defined by:
\begin{equation}\label{eq:8}
|\nabla_wg(u)|=\sqrt{\sum_{i=1}^m{|\nabla_wg_i(u)|^2}}
\end{equation}
with
\begin{equation}\label{eq:9}
| \nabla_wg_i(u)|=\sqrt{\sum_{v\not =u}w(u,v)(g_i(u)-g_i(v))^2} \qquad i=1,2,...,m
\end{equation}
where \(g_i(u)\) is the \(i^\mathrm{th}\) component of \(g(u)\), which have \(m\) components.
Using the above definitions we can regularize the function \(g\) using the following algorithm:
\begin{align}\label{eq:5}
& g_i^{(0)}=g_i & \\
& g_i^{(t+1)}(v)=\frac{\lambda g_{i}^{(0)}(v)+\sum_{u\not=v} \gamma_i^{(t)}(u,v)g_i^{(t)}(u)}{\lambda+\sum_{u\not=v}\gamma_i^{(t)}(u,v)} &
\end{align}
with 
\begin{equation}\label{eq:10}
\gamma_i^{(t)}(u,v)=w(u,v)(|\nabla_wg_i^{(t)}(u)|^{p-2}+|\nabla_wg_i^{(t)}(v)|^{p-2})
\end{equation}
where the parameter \(p\) controls the degree of regularity which has to be preserved and the parameter \(\lambda\) controls the fidelity to the original function \(g^0\).
\subsubsection{Detecting Adversarial Examples using the Regularization of the feature vector}\label{sec:3.1.2}
\indent
\par For this method we use a set of input images \(S\) that consists of real images and adversarial images, for which we know the correct target categories \(C_s\). Also, we have a set of new images \(V\) for which we do not know whether there are adversarial or not. For the images of both  sets \(S\) and \(V\)  we get the feature vectors that are extracted from the trained CNN and we create a weighted graph. Using this graph we perform the regularization that was described in  section \ref{sec:3.1.1} using \(p=1\). Then we retrain the last layer of the network, which takes as input the feature vectors, using as inputs the regularized feature vectors  only from the images of the  \(S\)  set and as desired outputs the  correct categories \(C_s\). A new image \(\bm{x} \in V\)   is detected as an adversarial image when the category into which is classified from the retrained last layer, using its regularized feature vector, is different from the category into which is classified by the original CNN. This  procedure is shown in Algorithm \ref{alg:1}.
\par An important detail of this method is the weights of the edges of the graph that are created from the input images. The weight of an edge which connects two images can depend on either  the distance between them, or the distance between their feature vectors.
\par Let \(g(u_1)\) be the feature vector of  image \(\bm{x}_1\) and \(g(u_2)\) be the feature vector of  image \(\bm{x}_2\). The weight of the edge between the vertices \(u_1,u_2\) using the distance of the feature vectors can be computed by:

\begin{equation}\label{eq:11}
w(u_1,u_2)=\exp{(-\frac{\lVert g(u_1)-g(u_2) \rVert_2^2}{\sigma^2})}
\end{equation}
When we use the distance between the two images, the weight of the edge between the vertices \(u_1,u_2\) can be computed by:
\begin{equation}\label{eq:12}
w(u_1,u_2)=\exp{(-\frac{\lVert \bm{x}_1-\bm{x}_2 \rVert_2^2}{\sigma^2})}
\end{equation}
\par The use of the Euclidean distance between the input images can be explained by the fact that we want the weights to illustrate the similarity between the images. However, the Euclidean distance expresses this similarity only when the input patterns are aligned and they are on the same scale. In contrast, the Euclidean distance between the feature vectors can express this similarity even when the input patterns are not aligned and have different scales, but with the drawback that this distance can be more easily manipulated by adversarial inputs. 
\par In addition to the Euclidean distance we can try to use different distances to compute the weights of the graph. So we may also use one of the following distances \(d\):
\begin{itemize}
\item the cosine distance: \(d(\bm{x_1},\bm{x_2})=1-\frac{\langle \bm{x_1},\bm{x_2}\rangle}{\lVert \bm{x_1} \rVert_2 \lVert \bm{x_2} \rVert_2}\).\\ (where \(\langle \bm{x}_1,\bm{x}_2\rangle\) is the inner product of \(\bm{x}_1\), \(\bm{x}_2\)  )
\item the \(L_1\) distance of the two vectors: \(d(\bm{x_1},\bm{x_2})=\rVert \bm{x_1}-\bm{x_2} \lVert_1\).
\end{itemize}
In order to use the distance function \(d\),  Equation (\ref{eq:11}) can be generalized as follows:
\begin{equation}\label{eq:13}
w(u_1,u_2)=\exp{(-\frac{d( g(u_1),g(u_2) )^2}{\sigma^2})}
\end{equation}
Also  Equation (\ref{eq:12}) can be generalized as follows:
\begin{equation}\label{eq:14}
w(u_1,u_2)=\exp{(-\frac{d( \bm{x}_1,\bm{x}_2 )^2}{\sigma^2})}
\end{equation}
Using  Equations (\ref{eq:13}), (\ref{eq:14}) with the different distance functions \(d\) we get different results of detecting adversarial examples. 

\begin{algorithm}[h]
\SetAlgoLined
\KwIn{Set \(S\) of images with known correct categories \(C_s\), Set V of images from which we want to detect the adversarial examples }
\DontPrintSemicolon
Compute the feature vectors \(h(S),h(V)\) for the images in the sets \(S,V\).\;
Using \(h(S),h(V)\) and one of the Equations (\ref{eq:13}),(\ref{eq:14}) create graph G.\;
Regularize graph G using p=1, and get the regularized feature vectors \(h_r(S),h_r(V)\).\;
Retrain the last layer of the CNN using as input the feature vectors \(h_r(S)\) and as target output the categories \(C_s\).\;
\ForEach{  \(\bm{x} \in V\)}{
 Using the original last layer \(L\) and the original feature vector \(h(\bm{x})\), get the classification category \(c_{orig}=L(h(\bm{x}))\),\;
 Using the retrained last layer \(L_{rt}\) and the regularized feature vector \(h_r(\bm{x})\), get the classification category \(c_{reg}=L_{rt}(h_r(\bm{x}))\),\;
 \uIf{ \(c_{orig}\not=c_{reg}\)}{
    Detect \(\bm{x}\) as an adversarial image. \;
 }
 \Else {
   Detect \(\bm{x}\) as a real image. \;
 }
}
\caption{Generalized Algorithm for detecting adversarial examples using the regularization of the feature vectors}  \label{alg:1}
\end{algorithm}
\pagebreak
\subsubsection{Experiments using the Regularization Method}\label{sec:3.1.3}
\indent
\par We generate adversarial examples using the BIM method and the DeepFool method on a LeNet \cite{LBBH98} neural network that was trained on the MNIST dataset. From the test set of the MNIST dataset we use 2000 images to create 2000 adversarial examples using the BIM method and 2000 adversarial examples using the DeepFool method. 
\begin{table}[H]
\centering
\makebox[\textwidth][c]{
\begin{tabular}{|c|c|c|c|c|c|c|c|c|}
\hline
\multirow{2}{*}{}& \multicolumn{8}{c|}{Adversarial Detection}\\
\cline{2-9}
& \multicolumn{4}{c|}{Using  Equation (\ref{eq:13})} & \multicolumn{4}{c|}{Using  Equation (\ref{eq:14})} \\
\cline{2-9}
& \multicolumn{2}{c|}{BIM} & \multicolumn{2}{c|}{DeepFool} & \multicolumn{2}{c|}{BIM} & \multicolumn{2}{c|}{DeepFool} \\
\hline
Distance& Precision & Recall & Precision & Recall & Precision & Recall & Precision & Recal\\
\hline
\(L_2\) & 89.5\% & 66.2\% & 92.4\% & 83.7\%  & 90.2\% & 67.1\% & 93.2\% & 84.4\%\\ 
\hline
Cosine & 90.2\% & 66.8\% & 91.9\% & 84.2\% & 90.7\% & 68.9\% & 92.7\% & 84.9\%\\ 
\hline
\(L_1\) & 90\% & 65.7\% & 91.2\% & 84.1\% & 90.8\% & 66.1\% & 92.5\% & 84.6\% \\ 
\hline
&\multicolumn{8}{c|}{Adversarial Detection without Regularization}\\
\cline{2-9}
& \multicolumn{4}{c|}{BIM} & \multicolumn{4}{c|}{DeepFool} \\
\cline{2-9}
& \multicolumn{2}{c|}{ Precision=86\% } & \multicolumn{2}{c|}{ Recall=51\% } &  \multicolumn{2}{c|}{ Precision=88.3\% } & \multicolumn{2}{c|}{ Recall=69.2\% }\\ 
\hline
\end{tabular}
}
\\[-1ex]
\caption{Results of adversarial detection on the LeNet network using the Regularization Method, when different distances are used to compute the weights of the edges of the graph} \label{tab:1}
\end{table}
\par We then use the Regularization Method that was presented in section \ref{sec:3.1.2} in order  to detect the adversarial images  in a set of 2000 real and 2000 adversarial images, using the different distances and the different adversarial generation methods. Also, we try to detect the adversarial examples without using the regularization, which means that the method retrains the last layer using the original feature vectors. When we remove the regularization of the feature vectors, retraining the last layer is similar to adversarial training \cite{GSS14}. The results are shown in  Table \ref{tab:1}. Both the highest Precision and the highest Recall in the detection is achieved when  the weights are computed using  Equation (\ref{eq:14}).

\subsection{Histogram Method for Adversarial Detection}\label{sec:3.2}
\indent
\par
\begin{figure}[h]
\centering
\makebox[\textwidth][c]{
\subfloat[][]{
\includegraphics[width=0.36\textwidth,height=0.2\textheight]{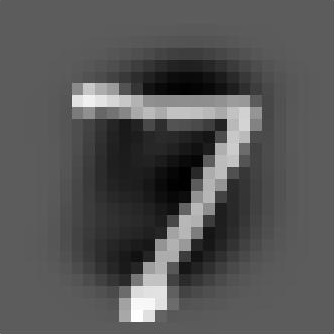}
\label{fig:7}
}
\subfloat[][]{
\includegraphics[width=0.4\textwidth,height=0.21\textheight]{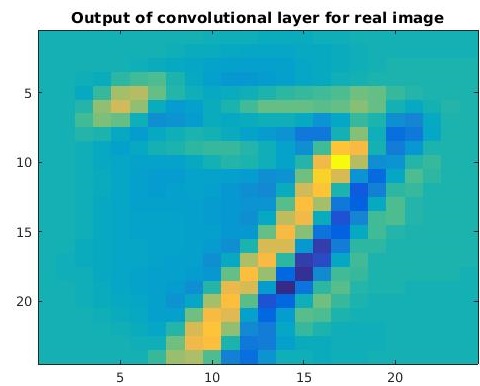}
\label{fig:8}

}
}
\\[-1ex]
\centering
\makebox[\textwidth][c]{
\subfloat[][]{
\includegraphics[width=0.36\textwidth,height=0.2\textheight]{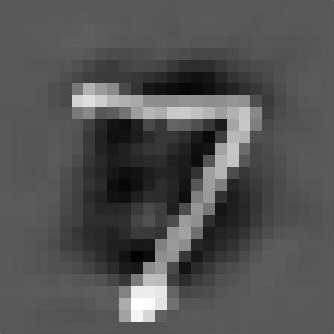}
\label{fig:9}
}
\subfloat[][]{
\includegraphics[width=0.4\textwidth,height=0.21\textheight]{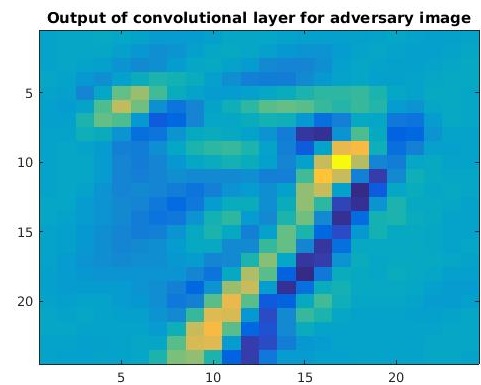}
\label{fig:10}
}
}

\caption{First column: Input images where \protect\subref{fig:7} is a real image and \protect\subref{fig:9} is an adversarial image. Second column: Output of the first convolutional layer of the neural network for the inputs of the first column. (Best viewed in color)}
\label{fig:102}

\end{figure}

 By comparing the outputs of the hidden layers of a  CNN when the inputs are real images and when the inputs are adversarial images, we can observe that the outputs in these two cases have different distribution of  values. In particular, we compare the outputs of an original image and an adversarial image generated from the original. An example of two such outputs is presented in Figure \ref{fig:102}. We observe that in the case of the adversarial image there is an increase in the values of some peaks of the original output while there is a decrease in the values on the rest of the points of the output.

\par
\begin{figure}[h]\center
\subfloat[][]{
\includegraphics[width=0.5\textwidth,height=0.3\textheight]{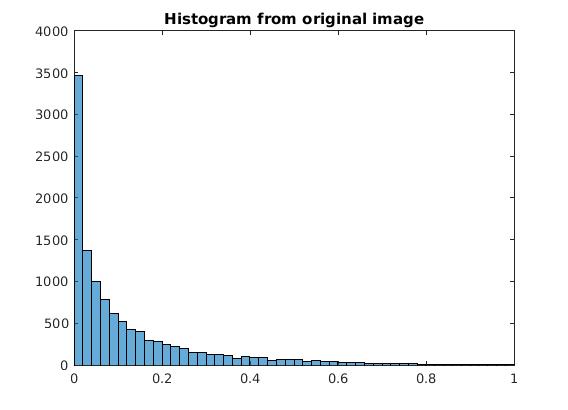}
\label{fig:11}
}
\subfloat[][]{
\includegraphics[width=0.5\textwidth,height=0.3\textheight]{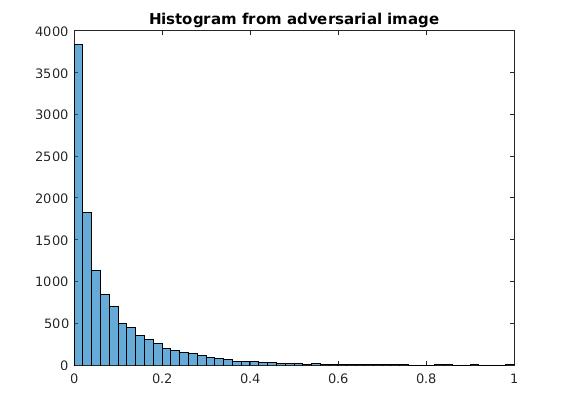}
\label{fig:12}
}
\caption{Histograms of the absolute values of the output when the input is: \protect\subref{fig:11} the real image of  Figure \ref{fig:7}, \protect\subref{fig:12} the adversarial image of  Figure \ref{fig:9}.}
\label{fig:103}
\end{figure}

 The difference in the distribution of the outputs can be detected using  histograms of the output values. In  Figure \ref{fig:103}, we can see histograms of the absolute values of the outputs of the first convolutional layer from the LeNet network, when the input is a real image and when  it is an adversarial image. In these histograms we can observe that in the case of the adversarial image, the points of the output, which have values with high absolute value, are less than the respective points in the case of the real image.
\par We can use this difference in the histograms for adversarial detection. To do so, we train an SVM classifier, which takes the histogram of the absolute values of the output of the first convolutional layer as input and predicts whether the input image is an adversarial image or a real image.
\par This method is problematic when we add Gaussian noise in the input images. The additive noise changes the distribution of the values of the outputs and as a result the output distribution when the inputs are adversarial images becomes more similar to the output distribution when the inputs are real images. To improve this method and make it more robust against the addition of Gaussian noise we propose the reinforcement step. Let \(f\) be the function that is implemented by the CNN to produce the network's final output  and \(\bm{x}\) be an input image that is classified into the category \(\ell\). Using the reinforcement step we get the new image \(\bm{x}_\mathrm{new}\) which is defined by:
\begin{equation}\label{eq:111}
\bm{x}_{new}=\bm{x}-\epsilon \frac{\nabla_xJ(f(\bm{x}),\ell)}{\lVert \nabla_xJ(f(\bm{x}),\ell)\rVert_2}\lVert \bm{x}\rVert_2
\end{equation}
where \(J(f(\bm{x}),\ell) \) is the classification error of the network for input \(\bm{x}\) when the target category is \(\ell\).
\par When the original input \(\bm{x}\) is an adversarial image the reinforcement step will increase the confidence for the adversarial category and as a result \(\bm{x}_\mathrm{new}\) will have a histogram which is more distinct from a histogram of a real image. Similarly, when the original input \(\bm{x}\) is a real image, \(\bm{x}_\mathrm{new}\) will increase the confidence for the real category and it will have a histogram which is more distinct from a histogram of an adversarial image. Hence, when we use both  \(\bm{x}_\mathrm{new}\) and  \(\bm{x}\) to produce the histograms that are used for detection, it is easier for the SVM classifier to distinguish between real and adversarial images.
\par Another detail that improves the detection is, when we create the histogram from the output of a layer which has \(n\) channels, to create one histogram for each channel instead of creating one histogram for all the channels. As a result, when we have a layer with \(n\) channels we get \(n\) different histograms and by combining them, we create the final vector that will be used as the  input of the SVM.

\begin{algorithm}[h]
\SetAlgoLined
\KwIn{Set \(\{\bm{x}_1,\bm{x}_2,...,\bm{x}_m\}\) of images which are a mix of real and adversarial images, set \(\{a_1,a_2,...,a_m\}\) where \(a_i=1\)  if \(\bm{x}_i\) is an adversarial image and \(a_i=0\) otherwise  }
\DontPrintSemicolon
\For{\(i\gets 1\) \KwTo \(m\)}
{
  Generate \(\bm{x}_{new}\) using Equation (\ref{eq:111}) and \(\bm{x}_i\),\;
  Compute the feature maps of the hidden layers for input \(\bm{x}_i\),\;
  \ForEach{ feature map  \(\bm{\mathrm{{fm}}}_j\)  of the first convolutional layer }
  {
    Compute the histogram \(\bm{h}_j\) of the absolute values of  \(\bm{\mathrm{{fm}}}_j\) ,\;
  }
  Concatenate the values of all \(\bm{h}_j\) to create \(\bm{hist}\)\;
  Compute the feature maps of the hidden layers for input \(\bm{x}_{new}\),\;
  \ForEach{ feature map \(\bm{\mathrm{{fm}}}_j\) of the first convolutional layer }
  {
    Compute the histogram \(\bm{h}'_j\) of the absolute values of  \(\bm{\mathrm{{fm}}}_j\) ,\;
  }
  Concatenate the values of all \(\bm{h}'_j\) to create \(\bm{hist}'\),\;
  Concatenate the values of \(\bm{hist},\bm{hist'}\) to create \(\bm{\mathrm{Thist}}_i\)
}
Train the SVM classifier using as inputs the \(\bm{\mathrm{Thist}}_i\) and as outputs the \(a_i\) for \(i=1,2,...,m\).\;
\caption{Training of the SVM that is used in the Histogram method that utilizes the reinforcement step} \label{alg:2}
\end{algorithm}
\subsubsection{Experiments using the Histogram Method} \label{sec:3.2.1}
\indent
\par Similarly with  section \ref{sec:3.1.3}, using a LeNet network trained on the MNIST dataset and both the BIM method and the DeepFool method we create two sets of 2000 adversarial and 2000 real images. We try to detect the adversarial images using both the original histogram method and the histogram method that utilizes the reinforcement step. The SVM that is used for the detection is trained from a different set of 1000 real and 1000 adversarial images generated using the BIM method. Also, we test the two methods when we add Gaussian noise with different values of standard deviation. 
\begin{table}[H]
\begin{center}
\begin{tabular}{|c|c|c|c|c|}
\hline
 \multicolumn{5}{|c|}{ Without Noise}\\
\hline
& \multicolumn{2}{c|}{BIM} & \multicolumn{2}{c|}{DeepFool} \\
\cline{2-5}
& Precision & Recall & Precision & Recall \\
\hline
Original & 97.9\% & 96.5\% & 98\% & 96.6\% \\ 
\hline
Reinforcement Step & 95.5\% & 94.6\%  & 95.4\% & 93.6\% \\ 
\hline
\multicolumn{5}{|c|}{ With Noise with standard deviation 15}\\
\hline
Original & 79.7\% & 74.9\%  & 80\% & 81.7\%\\ 
\hline
Reinforcement Step & 85.5\% & 87.2\% & 86.8\% & 90.7\% \\ 
\hline
\end{tabular}
\end{center}
\caption{Results of adversarial detection on the LeNet network using the original histogram method and the method that incorporates the reinforcement step, with and without additive Gaussian noise on the input.} \label{tab:2}
\end{table}
\par
\begin{figure}[h]
\center
\subfloat[][]{
\includegraphics[width=0.5\textwidth,height=0.31\textheight]{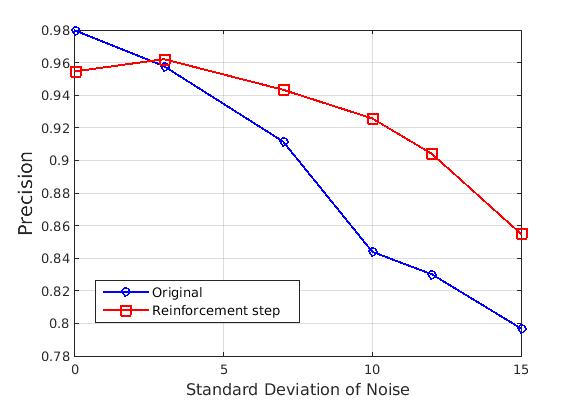}
\label{fig:13}
}
\subfloat[][]{
\includegraphics[width=0.5\textwidth,height=0.31\textheight]{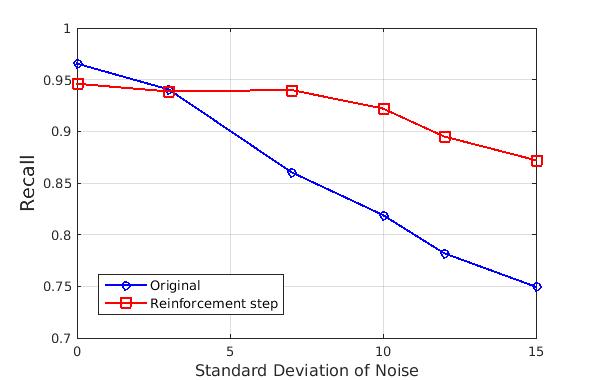}
\label{fig:14}
}
\caption{Adversarial detection results using the histogram methods for different values of standard deviation of the additive Gaussian noise: \protect\subref{fig:13} Precision of the detection \protect\subref{fig:14} Recall of the detection}

\label{fig:104}
\end{figure}

 The results of the experiments when there is no additive noise and when there is additive Gaussian noise with standard deviation 15 are presented in  Table \ref{tab:2}. Also  Figures \ref{fig:13}, \ref{fig:14}  show the Precision and the Recall of the two histogram methods for different values of the standard deviation of the additive Gaussian noise. When there is no additive noise the original method achieves  better results. Nevertheless, the difference between the methods when there is no noise is small, and because the method utilizing the reinforcement step is much more robust to the addition of noise, we can conclude that it is the preferable method.

\subsection{ Adversarial Detection using the Residual Image}\label{sec:3.3}
\indent
\par For the third proposed method we first introduce the concept of the residual image. Then, using the information of the residual image, we propose a way of  detecting whether the input image is an adversarial input by perturbing the input image.  
\par First, we propose a simple way of utilizing the residual image for  adversarial detection (Method A) and then we present two alternative methods (Methods B,C), that achieve better results in the detection of adversarial images on the MNIST dataset. 
\subsubsection{Residual Image}\label{sec:3.3.1}
\newcommand{\costumsd}{0.34\textwidth}
\begin{figure}[h]
\centering
\subfloat[][]{
\includegraphics[width=\costumsd,height=\costumsd]{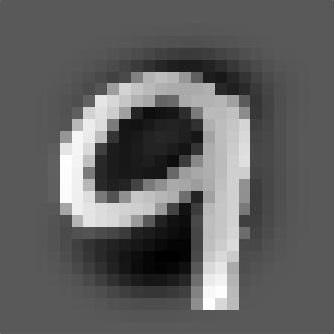}
\label{fig:15}
}
\subfloat[][]{
\includegraphics[width=\costumsd,height=\costumsd]{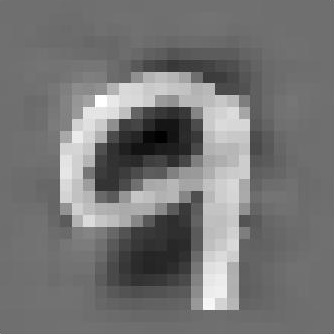}
\label{fig:16}
}
\\[-2ex]
\subfloat[][]{ 
\includegraphics[width=\costumsd,height=\costumsd]{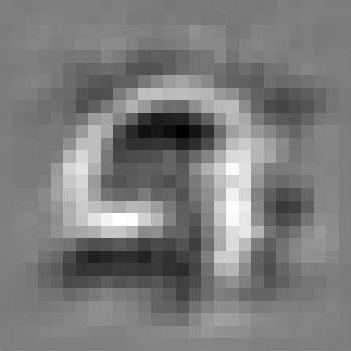}
\label{fig:17}
} 
\subfloat[][]{ 
\includegraphics[width=\costumsd,height=\costumsd]{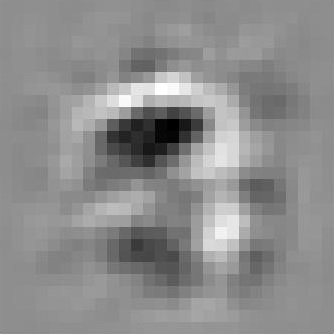}
\label{fig:18}
} 
\caption{\underline{First row}: Input images where image \protect\subref{fig:15} is a real input and image \protect\subref{fig:16} is an adversarial input. \underline{Second row}: \protect\subref{fig:17} Vector \(\bm{\delta}\) from Equation (\ref{eq:16}) when the input image is the real image. \protect\subref{fig:18} Vector \(\bm{\delta}\) from Equation (\ref{eq:16}) when the input image is the adversarial image.}
\label{fig:105}

\end{figure}
\indent
\par In a CNN the lower layers, closest to the input, act as feature extractors for the input image. The feature vector that is produced  is then used as an input for the layers closest to the output, which  are usually fully connected layers, in order  to classify the input into one of the possible categories.  Given an input image, we want to find an image related vector that if it is added to the input, it  will increase the norm of its feature vector  without changing its direction. Let \(h\) be the function that it is implemented by the lower layers and associates the input with a feature vector, \(\bm{x}_0\) be the original input and \(h(\bm{x}_0)=\bm{p}\) be the feature vector. We want to find a vector \(\bm{\delta} \) for which 
\begin{equation}\label{eq:16}
h(\bm{x}_0+\bm{\delta})=(1+\epsilon)\bm{p} \quad \epsilon>0 
\end{equation}
\par We can easily find the vector \(\bm{\delta}\) by using the backpropagation algorithm. As an example, by using a LeNet network trained on the MNIST dataset and the input images shown in Figures \ref{fig:15},\ref{fig:16}, we can compute the respective vectors \(\bm{\delta}\) which are shown in  Figures \ref{fig:17},\ref{fig:18}. We can see that for the input image of Figure \ref{fig:15}, which is a correctly classified real image, the vector \(\bm{\delta}\) resembles a /9/ digit, which is the correct category of the input image. In contrast, for the adversarial  input image of Figure \ref{fig:16}, which is classified falsely as a /7/ digit, the vector \(\bm{\delta}\) more closely resembles a pattern of the adversarial category than a pattern of the correct category. Therefore,  in a way the vector \(\bm{\delta}\) shows us the pattern that is perceived by the network.
\par This observation can be interpreted as follows: Due to the fact that the nonlinearities used in a CNN (e.g. Max Pooling, ReLU) are piecewise linear, in a neighborhood close to an input image the network acts as an affine classifier. This means that if we get the output \(h(\bm{x}_0)\) of the network, which is the feature vector of  image \(\bm{x}_0\), we can find a matrix \(\bm{W}\) and a vector \(\bm{b}\) so that 
\begin{equation}\label{eq:17}
h(\bm{x}_0)=\bm{W}\bm{x}_0+\bm{b}
\end{equation}
\par Given the input image \(\bm{x}_0\), the final layers of the neural network perceive that input as the feature vector \(\bm{h}_{x_0}=h(\bm{x}_0)\). Hence the neural network perceives the similarity between image \(\bm{x}_0\) and a new image \(\bm{x}\), which has feature vector \(\bm{h}_{x}\), as the \mbox{inner product  \(\bm{h}_{x_0}^T\bm{h}_x\)}. When these two inputs activate the nonlinearities of the neural network with the same way, we can use Equation (\ref{eq:17}) to compute the inner product  \(\bm{h}_{x_0}^T\bm{h}_x\) as follows: 
\begin{equation} \label{eq:18}
\bm{h}_{x_0}^T\bm{h}_{x}=\bm{h}_{x_0}^T\bm{W}\bm{x}+\bm{h}_{x_0}^T\bm{b}
\end{equation}
\par According to Equation (\ref{eq:18}) the similarity between the two images depends on the inner product of the new image \(\bm{x}\) with the image \(\bm{W}^T\bm{h}_{x_0}\). Hence the image \(\bm{W}^T\bm{h}_{x_0}\), which is produced by the backpropagation, can be interpreted as the pattern that is perceived by the network and it is used to find the similarity between the image \(\bm{x}_0\) and the new image \(\bm{x}\).
\par Let \(\bm{y}=[y_1,y_2,...,y_N]^T\) be the final output of the neural network, which classifies the input into \(N\) different categories, where \(y_i\) is the confidence of the network that the correct category of the  input is the category \(i\). Similarly with  Equation (\ref{eq:17}), in a neighborhood of the input space where the nonlinearities are activated with the same way, we can find a matrix \(\bm{A}\) and a vector \(\bm{d}\) so that for the input image \(\bm{x}\) the output \(\bm{y}=f(\bm{x})\) can be computed as follows:
\begin{equation} \label{eq:19}
\bm{y}=\begin{bmatrix} y_1 \\ y_2 \\ \vdots \\y_m \end{bmatrix}=\bm{A}\bm{x}+\bm{d}
\end{equation}
We can find which parts of the input pattern the neural network ignores by using the residual image \(\bm{x}_\mathrm{ign}\), which is the projection of the input image \(\bm{x}\) onto the null space of  matrix \(\bm{A}\). Also, we can get the perceived  image \(\bm{x}_\mathrm{p}=\bm{x}-\bm{x}_\mathrm{ign}\) which shows us the parts of the  pattern perceived by the network.
\newcommand{\costumsc}{0.25\textwidth}
\begin{figure}[h] 
\centering
\captionsetup[subfigure]{labelformat=empty}
\makebox[\textwidth][c]{
\subfloat[][]{
\includegraphics[width=\costumsc,height=\costumsc]{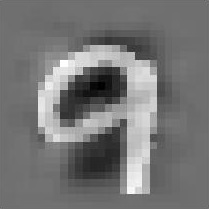}
\label{fig:19}
}
\subfloat[][]{
\includegraphics[width=\costumsc,height=\costumsc]{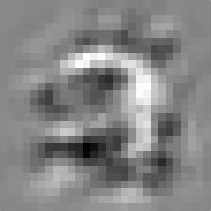}
\label{fig:20}
}
\subfloat[][]{
\includegraphics[width=\costumsc,height=\costumsc]{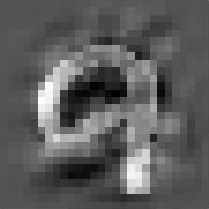}
\label{fig:21}
}
\subfloat[][]{
\includegraphics[width=\costumsc,height=\costumsc]{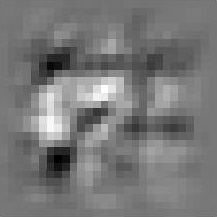}
\label{fig:22}
}
}
\\[-2ex]
\captionsetup[subfigure]{labelformat=empty}
\makebox[\textwidth][c]{
\subfloat[][(a)]{
\includegraphics[width=\costumsc,height=\costumsc]{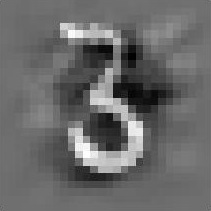}
\label{fig:23}
}
\subfloat[][(b)]{
\includegraphics[width=\costumsc,height=\costumsc]{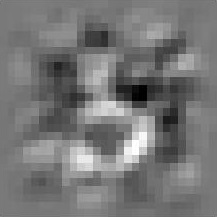}
\label{fig:24}
}
\subfloat[][(c)]{
\includegraphics[width=\costumsc,height=\costumsc]{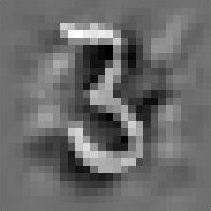}
\label{fig:25}
}
\subfloat[][(d)]{
\includegraphics[width=\costumsc,height=\costumsc]{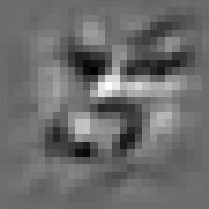}
\label{fig:26}
}
}
\caption{\underline{Column}: (a) Adversarial input images, (b) Perceived images \(\bm{x}_p\), (c) Residual images \(\bm{x}_\mathrm{ign}\), (d) Gradient of the classification error of the network when the input is  the adversarial image and the target category is the category the input is original classified into}
\end{figure}

\subsubsection{Detecting Adversarial Examples using the Residual Image}\label{sec:3.3.2}
\indent
\par Ideally a classifier, in order to classify a pattern of an image into one category must perceive the entire pattern. This means that in an ideal classifier we would expect that the norm of the residual image \(\bm{x}_\mathrm{ign}\) will be close to zero. In reality this does not happen and the norm of  \(\bm{x}_\mathrm{ign}\) stays high even for images of the training set. Nevertheless, what we can observe is that between a real image and a adversarial image generated by the real one, the norm of the ignored image \(\bm{x}_\mathrm{ign}\) increases as we go from the real image to the adversarial image. 

\par In the case of adversarial images, as we have shown in the section \ref{sec:3.3.1}, the perceived image \(\bm{x}_\mathrm{p}\)  resembles more the adversarial category. This means that the residual image \(\bm{x}_\mathrm{ign}\) contains valuable information for the distinction between the adversarial category and the real category, which is lost due to the fact that  the classifier perceives only the \(\bm{x}_\mathrm{p}\) image. The method of detecting adversarial examples is based on this observation and tries to add the lost information of the residual image \(\bm{x}_\mathrm{ign}\),  so that  it is not ignored by the classifier. In particular adding the information of the residual image into the original image, regardless if the original image is real or adversarial, will increase the confidence of the classifier for the correct category. As a result, in the case of an adversarial image, which is at first classified into the adversarial category, we will observe a decrease of the confidence for the adversarial category and an increase of the confidence for the correct category.
\par Let \(\bm{x}\) be the original image and \(\bm{x}_{\mathrm{ign}}\) be the residual image. Because in a neighborhood close to the image \(\bm{x}\) the output \(\bm{y}\) of the network can be computed using \mbox{Equation (\ref{eq:19})} and \(\bm{x}_{\mathrm{ign}}\) belongs to the null space of  matrix \(\bm{A}\), when we add \(\bm{x}_{\mathrm{ign}}\) to  \(\bm{x}\) the output of the network does not change.
\par If the input \(\bm{x}\) is classified by the network into the category \(\ell\), then if we add or subtract the gradient of the classification error \(\bm{\mathrm{grad}}=\nabla_xJ(f(\bm{x}),\ell)\), we can achieve the greatest change in the error of the classifier. Because the greatest change can be achieved either by adding or subtracting  \(\bm{\mathrm{grad}}\), we can use the information of the residual image \(\bm{x}_{\mathrm{ign}}\) to determine the direction that increases the confidence for the correct category. In the next sections we propose different methods, which can be used to  combine the information of \(\bm{x}_{\mathrm{ign}}\) and the information of \(\bm{\mathrm{grad}}\), in order to find the image \(\bm{x}_\mathrm{add}\) that we will add to the  image \(\bm{x}\) .

\subsubsection{Method A} \label{sec:3.3.3}
In the first method, at each pixel, we take the value of \(\bm{\mathrm{grad}}\) and the sign of \(\bm{x}_{\mathrm{ign}}\). The image \(\bm{x}_\mathrm{add}\) we want to add  is defined by:
\begin{align}
\bm{x}_\mathrm{r} &=R(\bm{x}_\mathrm{ign}) \label{eq:15} \\ 
\bm{x}_\mathrm{add} &=\mathrm{sign}(\bm{x}_\mathrm{r}\odot \bm{\mathrm{grad}})\odot \bm{\mathrm{grad}} \label{eq:116}
\end{align}
where \(R\) is a regularization function and  \(\odot\) is used to  refer to  the entrywise product of two vectors.
\par Although the regularization in Equation (\ref{eq:15}) is optional, the results from the experiments showed that by regularizing the residual image \(\bm{x}_\mathrm{ign}\) we can achieve an improvement in the adversarial detection. In the experiments presented in section \ref{sec:3.3.6} we use Total Variation Regularization \cite{ROF92}  in order to regularize the residual image \(\bm{x}_\mathrm{ign}\).
\par Method A starts with  an image \(\bm{x}_0\), which is classified into  category \(\ell\),  and perturbs it by iteratively adding \(\bm{x}_\mathrm{add}\) which is computed using  Equations (\ref{eq:15}),(\ref{eq:116}). After a certain number of iterations (\(T_\mathrm{max}\)) the image \(\bm{x}_\mathrm{final}\), that is produced, is used as an input to compute the  output \(\bm{s}=\mathrm{softmax}(f(\bm{x}_\mathrm{final}))\), which is the output of the  CNN after it goes through a softmax layer. Then the method detects the original image \(\bm{x}_0\) as an adversarial image if the softmax output for category \(\ell\)  is below a certain threshold \(\theta\). This procedure is shown in  Algorithm \ref{alg:3}. A weakness of this method is that with the  way it computes  \(\bm{x}_\mathrm{p}\), \(\bm{x}_\mathrm{ign}\), there is not a clear distinction between the parts of the pattern that are perceived and the parts that are ignored.  \\

\begin{algorithm}[H]
\SetAlgoLined
\KwIn{Image vector \(\bm{x}_0\) that we want to detect if it is an adversarial input}
\DontPrintSemicolon
Compute category \(\ell\) into which  \(\bm{x}_0\) is classified  from the CNN.\;
\(\bm{x}_\mathrm{cur}(0)\gets\bm{x}_0\).\;
\For{\(t\gets 0\) \KwTo \((T_{\mathrm{max}}-1)\) }{
 Compute \(\bm{x}_\mathrm{ign}\) for the input \(\bm{x}_\mathrm{cur}(t)\),\;
 Using \(\bm{x}_\mathrm{ign}\) and  Equations   (\ref{eq:15}),(\ref{eq:116}), compute \(\bm{x}_\mathrm{add}\),\;
 \(\bm{x}_\mathrm{cur}(t+1)\gets\bm{x}_\mathrm{cur}(t)+\epsilon \cdot \bm{x}_\mathrm{add}\),\;
}
\(s_\ell\gets\mathrm{softmax}(f(\bm{x}_\mathrm{cur}(T_\mathrm{max})))_\ell\) \quad (the softmax output  for the category \(\ell\))\;
If (\(s_\ell<\theta\))   detect \(\bm{x}_0\)  as an adversarial example
\caption{Method A}  \label{alg:3}
\end{algorithm}

\par

\subsubsection{Method B} \label{sec:3.3.4}
\indent
\par In order to make a clearer distinction between the parts of the pattern ignored and the parts perceived, this alternative method alters the way we compute the residual image. We denote this alternative residual image as \(\widetilde{\bm{x}}_\mathrm{ign}\).
\par  Using  Equation (\ref{eq:19}) we can find the projection of the input \(\bm{x}\) onto the null space of  matrix \(\bm{A}\) as follows:
\begin{equation}\label{eq:20}
\bm{x}_\mathrm{ign}=\bm{x}-\bm{A}^{\dagger} (\bm{y}-\bm{d})
\end{equation}
where \(\bm{A}^\dagger\) is the pseudoinverse of  matrix \(\bm{A}\)
\par With the alternative way of computing the residual image, we want the perceived image to be a clear depiction of a pattern that belongs to the category into which the input is classified. To achieve that, we want to change the output \(\bm{y}\) of the network and subsequently  using  Equation (\ref{eq:20}) to change the residual image. We find that in the case of images that belong to the training set there is a clearer distinction between perceived and ignored images. Therefore we want to make the output of the network to resemble the output  of the images of the training set.
\par To achieve that, we take the outputs of the training set and using the kmeans \mbox{algorithm} we find the centers of the clusters that these outputs create. Then given the original output \(\bm{y}\) we find the output \(\bm{y}_\mathrm{cent}\) which is the center closest to \(\bm{y}\). The alternative residual images \(\widetilde{\bm{x}}_\mathrm{ign}\) is defined by :
\begin{equation}\label{eq:21}
\widetilde{\bm{x}}_\mathrm{ign}=\bm{x}-\bm{A}^\dagger(\bm{y}_{\mathrm{cent}}-\bm{d})
\end{equation}
Hence, by replacing  \(\bm{x}_\mathrm{ign}\) with  \(\widetilde{\bm{x}}_\mathrm{ign}\) in the steps described in  Method A, we get the alternative Method B, which is presented in Algorithm \ref{alg:4}. This method improves the results of the detection, but produces an additive image \(\bm{x}_\mathrm{add}\) that is still noisy, which means that after a certain number of iterations the results of the detection using this method start to get worse.

\begin{algorithm}[H]
\SetAlgoLined
\KwIn{Image vector \(\bm{x}_0\), Set \(\{\bm{y}_{c_0},\bm{y}_{c_1},...,\bm{y}_{c_M}\}\) with the centers of the clusters of the outputs produced by the images of the training set}
\DontPrintSemicolon
Compute category \(\ell\) into which  \(\bm{x}_0\) is classified  from the CNN.\;
\(\bm{x}_\mathrm{cur}(0)\gets\bm{x}_0\).\;
\For{\(t\gets 0\) \KwTo \((T_{\mathrm{max}}-1)\) }{
 Compute  output \(\bm{y}\) for the input \(\bm{x}_\mathrm{cur}(t)\),\;
 Find center \(\bm{y}_\mathrm{cent}\) that is the closest to \(\bm{y}\),\;
 \(\widetilde{\bm{x}}_\mathrm{ign}\gets\bm{x}_{cur}(t)-\bm{A}^\dagger(\bm{y}_{\mathrm{cent}}-\bm{d})\),\;
 \(\bm{x}_\mathrm{add} \gets \mathrm{sign}(\widetilde{\bm{x}}_{ign}\odot \bm{\mathrm{grad}})\odot \bm{\mathrm{grad}}\),\;
 \(\bm{x}_\mathrm{cur}(t+1)\gets\bm{x}_\mathrm{cur}(t)+\epsilon \cdot \bm{x}_\mathrm{add}\),\;
}
\(s_\ell\gets\mathrm{softmax}(f(\bm{x}_\mathrm{cur}(T_\mathrm{max})))_\ell\) \quad (the softmax output  for the category \(\ell\))\;
If (\(s_\ell<\theta\))   detect \(\bm{x}_0\)  as an adversarial example
\caption{Method B}  \label{alg:4}
\end{algorithm}
\subsubsection{Method C} \label{sec:3.3.5}
\indent
\par
\begin{figure}[H]
\centering
\subfloat[][]{
\includegraphics[scale=0.84]{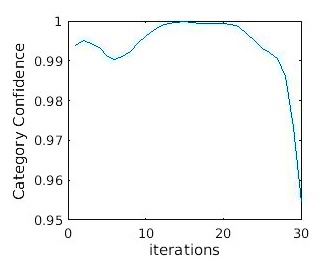}
\label{fig:27}
}
\subfloat[][]{
\includegraphics[scale=0.84]{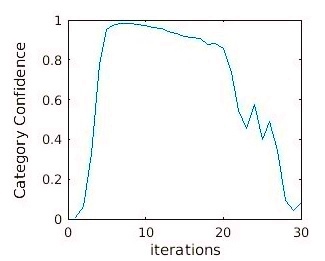}
\label{fig:28}
}
\\[-2ex]
\caption{Confidence of the classifier about the correct category when  Method B is used: \protect\subref{fig:19} for a real image, \protect\subref{fig:20} for an adversarial image}
\label{fig:107}
\end{figure}

 The iterative methods A,B, presented in the previous sections, do not converge to a final image and after a certain number of iterations they start to diverge. This is illustrated in  Figure \ref{fig:107}, where although at the first few iterations the image produced by  method B is classified into the correct category for both the real and the adversarial image, after a certain number of iterations the confidence of the network about the correct category starts to decrease.
\par To solve this problem, we change the way  we compute both the residual image \(\bm{x}_\mathrm{ign}\) and the image \(\bm{x}_\mathrm{add}\) that we  add during the method. Let \(\bm{x}_0\) be an original input image, \(\bm{x}_\mathrm{cur}\) be an image that is produced by the method from the original image \(\bm{x}_0\), and \(\bm{x}_\mathrm{p}\) be the perceived image when the input is the \(\bm{x}_\mathrm{cur}\) image. The alternative images \(\hat{\bm{x}}_\mathrm{ign}\),\(\hat{\bm{x}}_\mathrm{add}\) are defined by:

\begin{equation} \label{eq:22}
\hat{\bm{x}}_\mathrm{ign}=\frac{(\bm{x}_{0} \odot \bm{x}_\mathrm{cur}) - (\bm{x}_{0} \odot \bm{x}_\mathrm{p})}{\lVert \bm{x}_{0} \rVert_2}
\end{equation}
\begin{equation} \label{eq:23}
\hat{\bm{x}}_\mathrm{add}=\lvert \hat{\bm{x}}_\mathrm{ign}\odot \bm{\mathrm{grad}} \rvert \odot \mathrm{sign}(\hat{\bm{x}}_\mathrm{ign})
\end{equation}
where \(\odot\) is used to refer to the entrywise product of two vectors.
\par These equations  emphasize  the differences between \(\bm{x}_\mathrm{cur}\) and  \(\bm{x}_\mathrm{p}\), at the points where the absolute values of the original image are high.
\par Another detail that improves the results of the detection is to confine the values of the input image to a certain range, by setting a minimum value \(u_\mathrm{min}\) and a maximum value \(u_\mathrm{max}\). Let \(x_\mathrm{cur}(t)[i]\) be the \(i^{th}\)  pixel of the image \(\bm{x}_\mathrm{cur}(t)\), then:
\begin{equation}\label{eq:24}
x_\mathrm{cur}(t+1)[i]=\max(\min(x_\mathrm{cur}(t)[i]+\epsilon \cdot \hat{x}_\mathrm{add}[i],v_\mathrm{max}),v_\mathrm{min})
\end{equation} 
\par
\begin{figure}[H]
\centering
\subfloat[][]{
\includegraphics[scale=0.84]{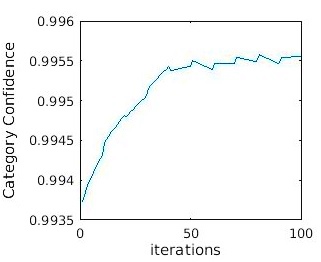}
\label{fig:29}
}
\subfloat[][]{
\includegraphics[scale=0.84]{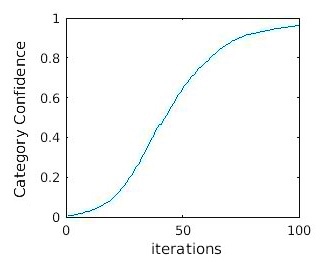}
\label{fig:30}
}
\\[-2ex]
\caption{Confidence of the classifier about the correct category when  Method C is used: \protect\subref{fig:21} for a real image, \protect\subref{fig:22} for an adversarial image}
\label{fig:108}
\end{figure}

 By replacing  \({\bm{x}}_\mathrm{ign}\),\({\bm{x}}_\mathrm{add}\) with  \(\hat{\bm{x}}_\mathrm{ign}\),\(\hat{\bm{x}}_\mathrm{add}\) in the steps described in  Method A and changing the way we compute \(\bm{x}_\mathrm{cur}(t+1)\) according to  Equation (\ref{eq:24}), we get the alternative Method C, which is presented  in Algorithm \ref{alg:5}. We can see how this method improves the convergence in   Figure \ref{fig:108}, where in contrast to the Method B, by increasing the iterations we do not observe a decrease in the confidence for the correct category.

\begin{algorithm}[H]
\SetAlgoLined
\KwIn{Image vector \(\bm{x}_0\), minimum value \(u_\mathrm{min}\) and  maximum value \(u_\mathrm{max}\) }
\DontPrintSemicolon
Compute category \(\ell\) into which  \(\bm{x}_0\) is classified  from the CNN.\;
\(\bm{x}_\mathrm{cur}(0)\gets\bm{x}_0\).\;
\For{\(t\gets 0\) \KwTo \((T_{\mathrm{max}}-1)\) }{
 Compute \(\bm{x}_\mathrm{p}\) for the input \(\bm{x}_\mathrm{cur}(t)\),\;
 Using \(\bm{x}_\mathrm{p}\), \(\bm{x}_\mathrm{cur}(t)\) and  Equations (\ref{eq:22}), (\ref{eq:23}), compute \(\hat{\bm{x}}_\mathrm{add}\),\;
 Update \(\bm{x}_\mathrm{cur}(t+1)\) , using Equation (\ref{eq:24})  .\;
}
\(s_\ell\gets\mathrm{softmax}(f(\bm{x}_\mathrm{cur}(T_\mathrm{max})))_\ell\) \quad (the softmax output  for the category \(\ell\))\;
If (\(s_\ell<\theta\))   detect \(\bm{x}_0\)  as an adversarial example
\caption{Method C}  \label{alg:5}
\end{algorithm}
\newcommand{\costumsw}{0.23\textwidth}
\begin{figure}
\captionsetup[subfigure]{labelformat=empty}
\centering
\subfloat{
\includegraphics[width=\costumsw,height=\costumsw]{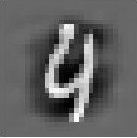}
\label{fig:31}
}
\subfloat{
\includegraphics[width=\costumsw,height=\costumsw]{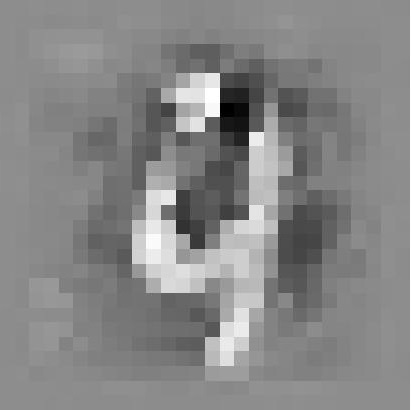}
\label{fig:32}
}
\subfloat{
\includegraphics[width=\costumsw,height=\costumsw]{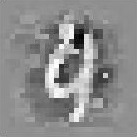}
\label{fig:33}
}
\subfloat{
\includegraphics[width=\costumsw,height=\costumsw]{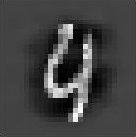}
\label{fig:34}
}

\subfloat[(a)]{
\includegraphics[width=\costumsw,height=\costumsw]{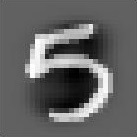}
\label{fig:35}
}
\subfloat[(b)]{
\includegraphics[width=\costumsw,height=\costumsw]{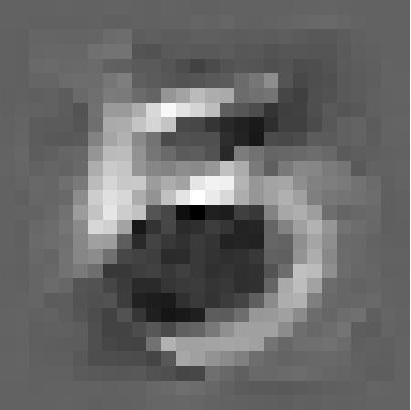}
\label{fig:36}
}
\subfloat[(c)]{
\includegraphics[width=\costumsw,height=\costumsw]{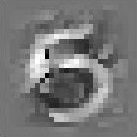}
\label{fig:37}
}
\subfloat[(d)]{
\includegraphics[width=\costumsw,height=\costumsw]{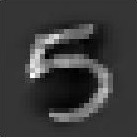}
\label{fig:38}
}
\caption{Comparison between the visual results of  Methods A,B,C: (a) Original input images, (b) Images produced after the termination of  Method A, (c)  Images produced after the termination of  Method B, (d)  Images produced after the termination of  Method C}
\label{fig:92}
\end{figure}

\subsubsection{Experiments with the proposed methods that use the residual image} \label{sec:3.3.6}
\indent
\par We use the three different methods that were presented in sections \ref{sec:3.3.3}, \ref{sec:3.3.4}, \ref{sec:3.3.5}  in order to detect the adversarial images in two sets of images, where each set contains 2000 real images and 2000 adversarial images. The experiments are  using the LeNet  network that is trained on the MNIST dataset and similarly with sections \ref{sec:3.1.3}, \ref{sec:3.2.1} the adversarial images for each set are generated using either the BIM method or the DeepFool method.
\par The results of the three methods using the residual image are presented in  \mbox{Table \ref{tab:3}}. It is clearly  shown  how the two alternative methods B,C improve the overall results of the detection when we compare them with the original method A. Also, it is worth noticing the fact that all the methods achieve much higher Recall, when they try to detect adversarial examples generated using the DeepFool method, compared to the Recall they achieve when they detect adversarial examples generated using the BIM method. This difference indicates that the adversarial examples generated from the DeepFool method are more sensitive to the perturbations applied by  Methods A,B,C, and as a result,  it is easier to enhance the correct category when the input image is an adversarial example generated from this method. 

\par One parameter that can greatly affect the results of these three methods is the threshold value \(\theta\) presented in Algorithms \ref{alg:3}, \ref{alg:4}, \ref{alg:5}. The detection results presented in Table \ref{tab:3} use the following threshold values:
\begin{itemize}
\item Method A: \(\theta=0.9\)
\item Method B: \(\theta=0.65\)
\item Method C: \(\theta=0.7\)
\end{itemize} 
\begin{table}[h]
\begin{center}
\begin{tabular}{|c|c|c|c|c|}
\hline
\multicolumn{5}{|c|}{Adversarial Detection}\\
\hline
& \multicolumn{2}{c|}{BIM} & \multicolumn{2}{c|}{DeepFool} \\
\cline{2-5}
& Precision & Recall & Precision & Recall \\
\hline
Method A & 68\% & 70.1\% & 73.5\% & 91.5\% \\ 
\hline
Method B & 84.6\% & 81.8\% & 86.7\% & 95.4\% \\ 
\hline
Method C & 87.6\% & 87.9\%  & 88.4\% & 94.7\%\\ 
\hline
\end{tabular}
\end{center}
\caption{ Adversarial detection results on the LeNet network using methods A,B,C.} 
\label{tab:3}
\end{table}
\begin{figure}[h]
\center
\makebox[\textwidth][c]{
\subfloat[][]{
\includegraphics[width=0.5\textwidth,height=0.32\textheight]{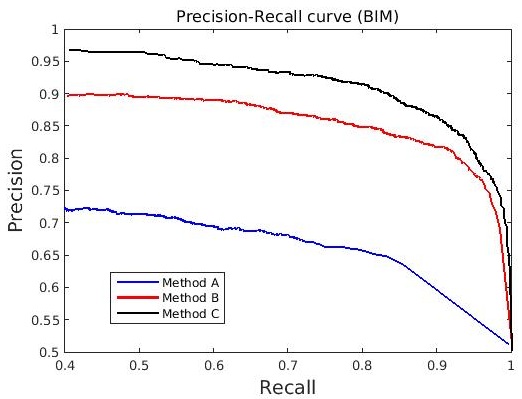}
\label{fig:39}
}
\subfloat[][]{
\includegraphics[width=0.5\textwidth,height=0.32\textheight]{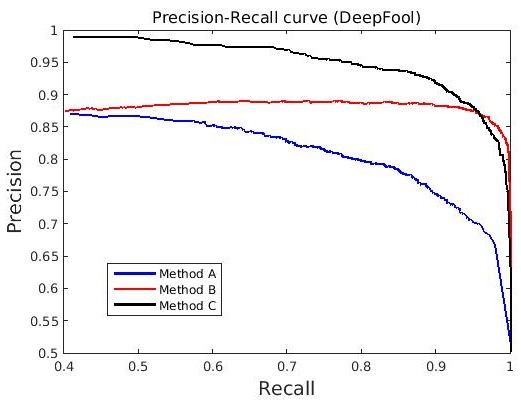}
\label{fig:40}
}
}
\caption{Precision-Recall curves of adversarial detection using Methods A,B,C when the adversarial examples are generated using: \protect\subref{fig:39} the BIM method, \protect\subref{fig:40} the DeepFool method. (Best viewed in color)}
\end{figure}
\par By changing the threshold value, we can change the Precision and the Recall of the methods according to the needs of each application. Generally, when we increase the threshold value, the detection becomes more sensitive and as a result the Precision increases, but at the same time the Recall decreases. In contrast, when we decrease the threshold value we observe a decrease in Precision and an increase in Recall. Therefore the best threshold value for the adversarial detection depends on how much each application values precision over recall. Figures \ref{fig:39}, \ref{fig:40} present the precision-recall curves for methods A,B,C , where we can observe the relationship between Precision and Recall for different threshold values. 
\par In  section \ref{sec:4}, where we  combine all the proposed methods, we use the same threshold values that were used to generate the results shown in Table \ref{tab:3}.
\section{Combining the results from the proposed methods} \label{sec:4}
\begin{figure}[h]
\centering
\makebox[\textwidth][c]{
\subfloat[][]{
\includegraphics[scale=0.6]{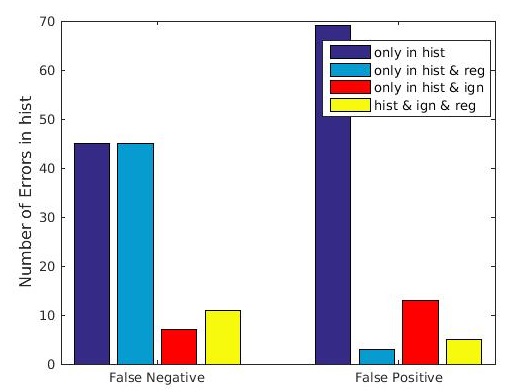}
\label{fig:41}
}
\subfloat[][]{
\includegraphics[scale=0.6]{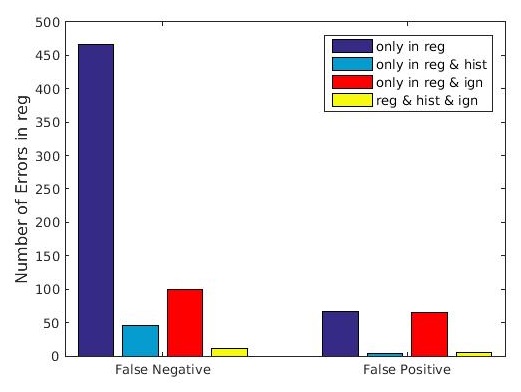}
\label{fig:42}
}
}

\makebox[\textwidth][c]{
\subfloat[][]{
\includegraphics[scale=0.6]{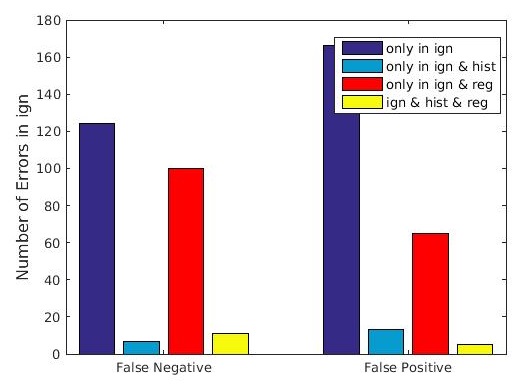}
\label{fig:43}
}
}
\caption{ Common errors between the proposed methods: histogram method(hist), regularization method(reg), residual image method(ign). (Best viewed in color)}

\label{fig:106}
\end{figure}
\indent
\par In order to  combine the results of the methods described in  sections \ref{sec:3.1}, \ref{sec:3.2}, \ref{sec:3.3}, it is useful to examine the common mistakes made from these methods. Hence we use the same set of 2000 real and 2000 adversarial images  and we try to detect the adversarial examples using each one of these methods. For the regularization method the weights are computed using the cosine distance between the input images, according to Equation (\ref{eq:14}), for the histogram method the reinforcement step is used and for the method utilizing the residual image  Method C is used. The results are illustrated in  Figure \ref{fig:106}.
\par These results show that each one of the three methods has a percentage of mistakes which are unique to the method. Another useful observation is that although the histogram method has the lowest number of false positive mistakes, where real images are detected as adversarial images, the majority of these falsely detected images are not detected as adversarial images by the other two methods. In contrast the other two methods have a large number of common false positive mistakes.
\par Let \(R(\bm{x})\) be the result of the regularization method, which takes the boolean value of 1 if the input image \(\bm{x}\) is detected as an adversarial image by the method. Similarly, let \(H(\bm{x})\) be the result of the histogram method, \(I(\bm{x})\) the result of the method that uses the residual image and \(T(\bm{x})\) the final result of detection after the combination of the three methods. Also we are using the symbol \(\lor\) for the logical OR and the symbol \(\land\) for the logical AND. Firstly, we identify an image as adversarial when it is detected by all the methods, which means that \(T_1(\bm{x})=(R(\bm{x}) \land H(\bm{x})\land I(\bm{x}))\). In this case, we expect to achieve the highest Precision but also the lowest Recall. In contrast if we identify an image as adversarial when it is detected by at least one method, which means that \(T_2(\bm{x})=(R(\bm{x}) \lor H(\bm{x}) \lor I(\bm{x}))\), then we will have the highest Recall with the lowest Precision. In addition if we identify an image as adversarial when it is detected by at least 2 methods , 
\(T_3(\bm{x})=((H(\bm{x}) \land R(\bm{x})) \lor (H(\bm{x}) \land I(\bm{x})) \lor (R(\bm{x}) \land I(\bm{x}))\), then we have intermediate results for both the Precision and the Recall.
\par In addition, in order to achieve high Precision without substantially decreasing the Recall, we can use the observation that the majority of the False Positive errors of the histogram method are unique to the method. So we identify an image as adversarial when it is detected as adversarial by the histogram method and by at least one of the other two methods, which means that \(T_4(\bm{x})=(H(\bm{x}) \land (R(\bm{x}) \lor I(\bm{x})))\).
The results of these combinations are shown in  Table \ref{tab:4}.
\begin{table}[H]
\begin{center}
\begin{tabular}{|c|c|c|c|c|}
\hline
\multicolumn{5}{|c|}{Result from the combination of the methods}\\
\hline
& \multicolumn{2}{|c}{BIM} & \multicolumn{2}{|c|}{DeepFool}\\
\hline
\(T\) & Precision & Recall  & Precision & Recall\\
\hline
\(T_1=R\land H \land I\) & 99.5\% & 60.1\% & 99.6\% & 77.2\% \\ 
\hline
\(T_2=R \lor H \lor I\) & 83.6\% & 99.4\% & 83.7\% & 99.7\% \\ 
\hline
\(T_3=((R\land H) \lor (R\land I) \lor (H\land I))\) & 95.5\% & 91.8\%  & 96.2\% & 96.25\%\\ 
\hline
\(T_4=H\land(R\lor I)\) & 98.8\% & 89.6\% & 98.9\% & 92.7\% \\ 
\hline
\end{tabular}
\end{center}
\caption{ Results of adversarial image detection on the LeNet network using different combinations  of the proposed  methods.}
\label{tab:4}
\end{table}
\par
\begin{figure}[H]
\center
\makebox[\textwidth][c]{
\subfloat[][]{
\includegraphics[width=0.5\textwidth,height=0.32\textheight]{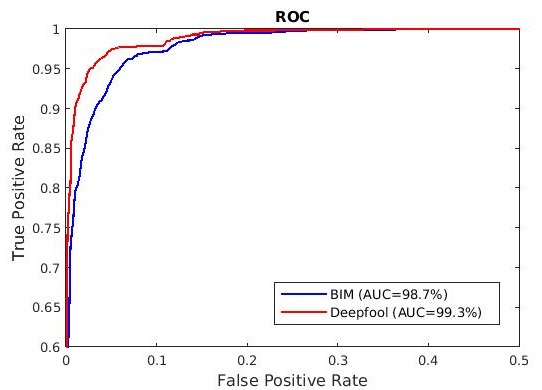}
\label{fig:201}
}
\subfloat[][]{
\includegraphics[width=0.5\textwidth,height=0.32\textheight]{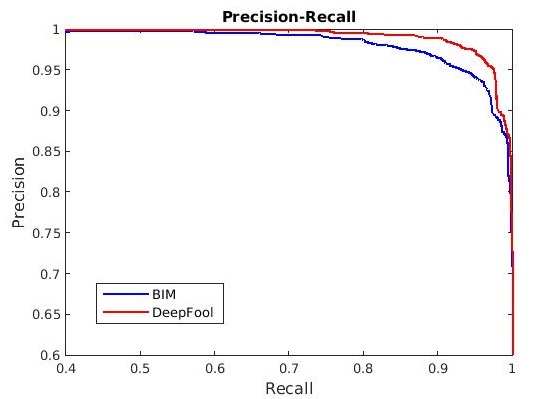}
\label{fig:202}
}
}
\caption{\protect\subref{fig:201} ROC curves of adversarial image detection, \protect\subref{fig:202} Precision-Recall curves,  which are generated by using different combinations of the proposed methods and different threshold values \(\theta\) for the method that utilizes the residual image. (Best viewed in color)}
\end{figure}

 Finally, we can compute the detection results of the combination of the proposed methods when we allow the threshold value  \(\theta\), of the method that utilizes the residual image, to change. If we use different values for \(\theta\) and we compute the detection results using the combinations \(T_1\), \(T_2\), \(T_3\) , we can create a ROC curve, which illustrates how we can change the sensitivity of the detection by changing both the threshold value \(\theta\) and the way we combine the three proposed methods. The ROC curves, which show the detection results when  we try to detect adversarial images  produced by the BIM and the DeepFool method, are presented in Figure \(\ref{fig:201}\). In addition, the respective \mbox{Precision-Recall} curves are presented in Figure \(\ref{fig:202}\) 
\par For the ROC curves the Area Under  Curve (AUC) is \(\mathrm{AUC}_{\mathrm{BIM}}=98.7\%\)  when we detect adversarial images that are generated using the BIM method, and \(\mathrm{AUC}_{\mathrm{DeepFool}}=99.3\%\) when we detect adversarial images that are generated using the DeepFool method. These results of adversarial image detection on the MNIST dataset compare favorably to the results of similar state of the art detection methods \cite{FCSG17}, \cite{Son+17}, which were briefly presented in section \ref{sec:2.2} and use a combination of metrics in order to detect adversarial images.
\section{Conclusion} \label{sec:5}
\indent
\par In this paper, we introduced three methods for detecting adversarial inputs in a CNN. Each one of these methods has some novelties and their combination yields an even more robust approach. The first method is based on adversarial retraining of the last layer of the network, and uses regularization of the input of the last layer to increase the effectiveness of the retraining. The second method uses the histograms of the values of the outputs of the hidden layers of the network in order to detect the adversarial inputs. Finally, in the third method we introduced the residual image, which gives us information about the parts of the input pattern that are ignored by the classifier. Using this information we  perturb  the input image in order to reinforce the correct category, something that allows us to detect the adversarial images which are not originally classified into the correct category.
\par After comparing the results of each individual method we showed how the combination of these methods improves the overall detection. This combination produces promising results and offers some improvements over similar approaches, when it is used for the detection of adversarial images on the MNIST dataset.

\end{document}